# Indexed Languages and Unification Grammars[*]

Tore Burheim[†]

**Abstract**

Indexed languages are interesting in computational linguistics because they are the least class of languages in the Chomsky hierarchy that has not been shown not to be adequate to describe the string set of natural language sentences. We here define a class of unification grammars that exactly describe the class of indexed languages.

## 1 Introduction

The occurrence of purely syntactical cross-serial dependencies in Swiss-German shows that context-free grammars can not describe the string sets of natural language [Shi85]. The least class in the Chomsky hierarchy that can describe unlimited cross-serial dependencies is indexed grammars [Aho68]. Gazdar discuss in [Gaz88] the applicability of indexed grammars to natural languages, and show how they can be used to describe different syntactic structures. We are here going to study how we can describe the class of indexed languages with a unification grammar formalism. After defining indexed grammars and a simple unification grammar framework we show how we can define an equivalent unification grammar for any given indexed grammar. Two grammars are equivalent if they generate the same language. With this background we define a class of unification grammars and show that this class describes the class of indexed languages.

## 2 Indexed grammars

Indexed grammars is a grammar formalism with generative capacity between context-free grammars and context-sensitive grammars. Context-free grammars can not describe cross-serial dependencies due to the pumping lemma, while indexed grammars can. However, the class of languages generated by indexed grammars, –the indexed languages, is a proper subset of context-sensitive languages [Aho68].

Indexed grammars can be seen as a context-free grammar where we add a string –or stack, of indices to the nonterminal nodes in the phrase structure trees, or derivation trees as we will call them. Some production rules add an index to the beginning of the string, while the use of other production rules is dependent on the first index in the string. When such a production rule is applied the index of which it is dependent, is removed, and the rest of the index-string is kept by the daughter(s). In this way we may distribute information from one part of the derivation tree to another. The original definition of indexed grammars was given

---

[*]This work has been supported by grant 100437/410 from Norwegian Research Council.

[†]University of Bergen, Department of Informatics, N-5020 Bergen, Norway *and* University of the Saarland, Computational Linguistic, Postfach 1150, D-66041 Saarbrücken, Germany. Email: Tore.Burheim@ii.uib.no

by Aho [Aho68]. We are here using the definition used by Hopcroft and Ullman [HU79] with some minor notational variations:

**Definition 1** *An* INDEXED GRAMMAR *$G$ is a 5-tuple; $G = \langle N, T, I, P, S \rangle$ where*

$N$ *is a finite set of symbols, called* nonterminals*,*

$T$ *is a finite set of symbols, called* terminals*,*

$I$ *is a finite set of symbols, called* indices*,*

$P$ *is a finite set of ordered pairs, each on one of the forms $\langle A, Bf \rangle$, $\langle Af, \alpha \rangle$ or $\langle A, \alpha \rangle$ where $A$ and $B$ are nonterminal symbols in $N$, $\alpha$ is a finite string in $(N \cup T)^*$, and $f$ is an index in $I$. An element in $P$ is called a* production rule *and is written $A \to Bf$, $Af \to \alpha$ or $A \to \alpha$.*

$S$ *is a symbol in $N$, and is called the* start symbol*.*

*and such that $N$, $T$ and $I$ are pairwise disjoint.*

*An indexed grammar $G = \langle N, T, I, P, S \rangle$ is on* REDUCED FORM *if each production in $P$ is on one of the forms*

a) $A \to Bf$

b) $Af \to B$

c) $A \to BC$

d) $A \to t$

*where $A, B, C$ are in $N$, $f$ is in $I$, and $t$ is in $(T \cup \{\varepsilon\})$.*

Aho showed in his original paper [Aho68] that for every indexed grammar there exists an indexed grammar on reduced form which generates the same language.

To define constituent structures and derivation trees we are going to use tree domains: Let $\mathcal{N}_+$ be the set of all integers greater than zero. A *tree domain* $D$ is a set $D \subseteq \mathcal{N}_+^*$ of number strings so that if $x \in D$ then all prefixes of $x$ are also in $D$, and for all $i \in \mathcal{N}_+$ and $x \in \mathcal{N}_+^*$, if $xi \in D$ then $xj \in D$ for all $j$, $1 \leq j < i$. The *out degree* $d(x)$ of an element $x$ in a tree domain $D$ is the cardinality of the set $\{i \mid xi \in D, i \in \mathcal{N}_+\}$. The set of terminals of $D$ is $term(D) = \{x \mid x \in D, d(x) = 0\}$. The elements of a tree domain are totally ordered lexicographically as follows: $x \preceq y$ if $x$ is a prefix of $y$, or there exist strings $z, z', z'' \in \mathcal{N}_+^*$ and $i, j \in \mathcal{N}_+$ with $i < j$, such that $x = ziz'$ and $y = zjz''$. We also define that $x \prec y$ if $x \preceq y$ and $x \neq y$.[1]

A tree domain $D$ can be viewed as a tree graph in the following way: The elements of $D$ are the nodes in the tree, $\varepsilon$ is the root, and for every $x \in D$ the element $xi \in D$ is $x$'s child number $i$. A tree domain may be infinite, but we shall restrict attention to finite tree domains. A finite tree domain can also describe the topology of a derivation tree. This representation provides a name for every node in the derivation tree directly from the definition of a tree domain. Our definition of derivation trees for indexed grammars with the use of tree domains is based on Hayashi [Hay73]:

**Definition 2** *A* DERIVATION TREE *based on an indexed grammar $G = \langle N, T, I, P, S \rangle$ is a pair $\langle D, C_{\mathcal{I}} \rangle$ of a finite tree domain $D$ and a function $C_{\mathcal{I}} : D \to (NI^* \cup T \cup \{\varepsilon\})$ where*

i) $C_{\mathcal{I}}(\varepsilon) = S$

---

[1] See Gallier [Gal86] for more about tree domains.

ii) $C_\mathcal{I}(x) \in NI^*$ for every node $x$ in $D$ with $d(x) > 0$. Moreover if $C_\mathcal{I}(x) = A\gamma$ for $A \in N$ and $\gamma \in I^*$ and $C_\mathcal{I}(xi) = B_i\theta_i$ with $B_i \in (N \cup T \cup \{\varepsilon\})$ and $\theta_i \in I^*$ for every $i : 1 \leq i \leq d(x)$ then either

   a) $A \to B_1 f$ is a production rule in $P$ such that $d(x) = 1$, $f \in I$, and $\theta_1 = f\gamma$, or
   
   b) $Af \to B_1 \ldots B_{d(x)}$ is a production rule in $P$ such that $f \in I$ where $\gamma = f\gamma'$, and $\theta_i = \gamma'$ if $B_i \in N$ and $\theta_i = \varepsilon$ if $B_i \in (T \cup \{\varepsilon\})$, or
   
   c) $A \to B_1 \ldots B_{d(x)}$ is a production rule in $P$ such that $\theta_i = \gamma$ if $B_i \in N$ and $\theta_i = \varepsilon$ if $B_i \in (T \cup \{\varepsilon\})$.

iii) $C_\mathcal{I}(x) \in (T \cup \{\varepsilon\})$ for every node in $D$ with $d(x) = 0$,

The SYMBOL FUNCTION; $C_\mathcal{I}^{sym} : D \to (N \cup T)$, and the INDEX STRING FUNCTION; $C_\mathcal{I}^{idx} : D \to I^*$, are total functions on $D$ such that if $C_\mathcal{I}(x) = A\gamma$ where $A \in (N \cup T \cup \{\varepsilon\})$ and $\gamma \in I^*$ then $C_\mathcal{I}^{sym}(x) = A$ and $C_\mathcal{I}^{idx}(x) = \gamma$ for all $x \in D$.

The TERMINAL STRING of a derivation tree $\langle D, C_\mathcal{I} \rangle$ is the string $C_\mathcal{I}(x_1)\ldots C_\mathcal{I}(x_n)$ where $\{x_1, \ldots, x_n\} = term(D)$ and $x_i \prec x_{i+1}$ for all $i, 1 \leq i \leq n-1$.

We also define the LICENSE FUNCTION; $license : (D - term(D)) \to P$, such that if $A \to \alpha$ is a production rule according to a), b) or c) in ii) for a node $x$ in $D$, then $license(x) = A \to \alpha$.

Informally this is a traditional phrase structure tree. If we have a node with label $A\gamma$ where $A$ is a nonterminal symbol and $\gamma$ is a string of indices, and we use a production rule $A \to Bf$, then the node's only child gets the label $Bf\gamma$. If we instead use a production rule $A \to BC$ on the same node it gets two children labeled $B\gamma$ and $C\gamma$ respectively, or if we use a production rule $A \to t$ where $t$ is a terminal symbol, then we remove all the indices and the node's only child gets the label $t$. If we have a node labeled with $Af\gamma$, where $f$ is a index and we use a production rule $Af \to B$ then the node's only child gets the label $B\gamma$. We also see that the terminal string is a string in $T^*$ since $C_\mathcal{U}(x) \in (T \cup \{\varepsilon\})$ for all $x \in term(D)$.

**Definition 3** *A string $w$ is* GRAMMATICAL *with respect to an indexed grammar $G$ if and only if there exists a derivation tree based on $G$ with $w$ as the terminal string. The language generated by $G$, $L(G)$ is the set of all grammatical strings with respect to $G$.*

**Example 1** Let $G = \langle N, T, I, P, S \rangle$ be an indexed grammar where $T = \{a, b, c\}$ is the set of terminal symbols, $N = \{S, S', A, B, C\}$ is the set of nonterminal symbols, $I = \{f, g\}$ is the set of indices and $P$ is the least set containing the following production rules:

$$
\begin{array}{lll}
S \to S'f & Ag \to aA & Af \to a \\
S' \to S'g & Bg \to bB & Bf \to b \\
S' \to ABC & Cg \to cC & Cf \to c
\end{array}
$$

Figure 1 shows the derivation tree for the string *"aabbcc"* based on this grammar. The language $L(G)$ generated by this grammar is $\{a^n b^n c^n \mid n \geq 1\}$.

We close this presentation of indexed grammars by showing a simple technical observation that we will use in later proofs.

**Definition 4** *An indexed grammar $G = \langle N, T, I, P, S \rangle$ has a* MARKED INDEX-END *if and only if it has one and only one production rule where the start symbol occurs and this rule is on the form $S \to A\$$ where $A \in N$ and the index $\$$ does not occur in any other production rule.*

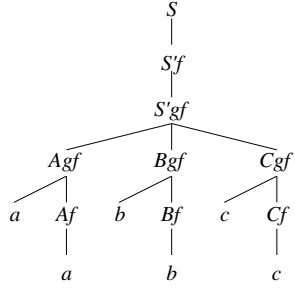

Figure 1: *Derivation tree for the string "aabbcc" based on the grammar in Example 1*

If an indexed grammar has a marked index-end then in any derivation tree every nonterminal node except the root gets a $ at the end of the index list. Since no rule requires that there is an empty index list, and neither $ nor the start symbol occurs in any other production rule, it is straight forward to construct an equivalent grammar with a marked index-end for any indexed grammar.

**Lemma 1** *For every indexed grammar $G$ there exists an indexed grammar with a marked index-end $G_\$$ such that $L(G) = L(G_\$)$.*

**Proof:** Let $G = \langle N, T, I, P, S \rangle$ be an indexed grammar, and assume that $S_0$ and $ do not occur in $G$. $G_\$$ is defined from $G$ by adding the production rule $S_0 \to S\$$ such that $S_0$ becomes the new start symbol and is added to the set of nonterminal symbols, and $ is added to the set of indices. Formally, if $G = \langle N, T, I, P, S \rangle$ and $S_0, \$ \notin (N \cup T \cup I)$, then $G_\$ = \langle N \cup \{S_0\}, T, I \cup \{\$\}, P \cup \{\langle S_0, S\$ \rangle\}, S_0 \rangle$. Then $G_\$$ has a marked index-end, and we have to show that for any string $w$, $w \in L(G)$ if and only if $w \in L(G_\$)$.

($\Longrightarrow$) Let $\langle D, C_\mathcal{I} \rangle$ be any derivation tree based on $G$ and assume that $w$ is its terminal string. From this we construct a derivation tree $\langle D', C'_\mathcal{I} \rangle$ based on $G_\$$ as follows: First let $D' = \{1x \mid x \in D\} \cup \{\varepsilon\}$. Then let $C'_\mathcal{I}(\varepsilon) = S_0$ and let $C'_\mathcal{I}(1x) = C_\mathcal{I}(x)\$$ for all $x \in (D - term(D))$. Let also $C'_\mathcal{I}(1x) = C_\mathcal{I}(x)$ for all $x \in term(D)$. The derivation tree $\langle D', C'_\mathcal{I} \rangle$ has then the same terminal string as $\langle D, C_\mathcal{I} \rangle$. Since no rule requires that there is an empty index list, and $ does not occur in any production rule in $G$, a production rule that is licensing a node $x$ in $\langle D, C_\mathcal{I} \rangle$, will license the node $1x$ in $\langle D', C'_\mathcal{I} \rangle$. The rule $S_0 \to S\$$ licenses the root. Then $\langle D', C'_\mathcal{I} \rangle$ is a valid derivation tree according to Definition 2.

($\Longleftarrow$) Let $\langle D', C'_\mathcal{I} \rangle$ be any derivation tree based on $G_\$$ and assume that $w$ is its terminal string. Since $S_0 \to S\$$ must license the root and $ does not occur in any other production rule the index symbol $ occurs at the end of the index list at every nonterminal node except the root in $\langle D', C'_\mathcal{I} \rangle$. From this derivation tree we construct a derivation tree $\langle D, C_\mathcal{I} \rangle$ based on $G$ as follows: First let $D = \{x \mid 1x \in D'\}$. Then for all $x \in (D - term(D))$ let $C_\mathcal{I}(x) = \beta$ where $C'_\mathcal{I}(1x) = \beta\$$. Let also $C_\mathcal{I}(x) = C'_\mathcal{I}(1x)$ for all $x \in term(D)$. The derivation tree $\langle D, C_\mathcal{I} \rangle$ has then the same terminal string as $\langle D', C'_\mathcal{I} \rangle$. Since every production rule in $G_\$$ except $S_0 \to S\$$ also is a production rule in $G$, the rule $S_0 \to S\$$ only can license the root, and $ does not occur in any other production rule, a production rule that licenses a node $1x$ in $\langle D', C'_\mathcal{I} \rangle$ will license the node $x$ in $\langle D, C_\mathcal{I} \rangle$. Then $\langle D, C_\mathcal{I} \rangle$ is a valid derivation tree according to Definition 2. □

Notice in the proof that if $G$ is on reduced form then $G_\$$ is also on reduced form. Then for any indexed grammar on reduced form there also exists an indexed grammar on reduced form with a marked index-end.

# 3 Unification grammars

We are here going to give a description of a very simple unification grammar formalism. The formalism itself is not particularly interesting, and it is only meant as a framework for the rest of this paper. The formalism is just a notational variant of the basic formalism used by Colban in his work on restrictions on unification grammars [Col91]. It should be easy to reformulate this in most of the known formalisms available. We give an informal description of feature structures in the way they are used here before we define the grammar formalism.

A *feature structure* over a set of attribute symbols $A$ and value symbols $V$ is a four-tuple $\langle Q, \delta, \alpha, m_D \rangle$ where $Q$ is a finite set of nodes, $\delta : Q \times A \to Q$ is a partial function, called the transition function, $\alpha : Q \to V$ is a partial function called the atomic value function, and $m_D : D \to Q$ is a function, called the name mapping. We will mostly omit the name-domain from the notation, so $m$ will alone denote the name mapping. We extend the transition function to be a function from pairs of nodes and *strings* of attribute symbols: For every $q \in Q$ let $\delta(q, \varepsilon) = q$. If $\delta(q_1, \psi) = q_2$ and $\delta(q_2, a) = q_3$ then let $\delta(q_1, \psi a) = q_3$ for every $q_1, q_2, q_3 \in Q$, $\psi \in A^*$ and $a \in A$.

A feature structure is *describable* if there for every node is a path from a named node to the node. This means that for every $q \in Q$ there is an $x \in D$ and a $\psi \in A^*$ such that $\delta(m(x), \psi) = q$. A feature structure is *atomic* if every node with an atomic value has no out-edges. This means that for every node $q \in Q$, $\delta(q, a)$ is not defined for any $a \in A$ if $\alpha(q)$ is defined. A feature structure is *acyclic* if it does not contain attribute cycles. This means that for every node $q \in Q$, $\delta(q, \psi) = q$ if and only if $\psi = \varepsilon$. A feature structure is *well defined* if it is describable, atomic and acyclic. When nothing else is said we require that feature structures are well defined in the rest of this paper.

We are going to use equations to describe feature structures, in a way where feature structure satisfies equations. A feature structure *satisfies* the equation

$$x_1 \psi_1 \doteq x_2 \psi_2 \tag{1}$$

if and only if $\delta(m(x_1), \psi_1) = \delta(m(x_2), \psi_2)$, and the equation

$$x_1 \psi_1 \doteq v \tag{2}$$

if and only if $\alpha(\delta(m(x_1), \psi_1)) = v$, where $x_1, x_2 \in D$, $\psi_1, \psi_2 \in A^*$ and $v \in V$. We only allow equations on those two forms. This means that there is no typing, quantification, implication, negation, or explicit disjunction as we may find in other unification grammars and feature logics.

If $E$ is a set of equations of the above form and $M$ is a well defined feature structure such that $M$ satisfies every equation in $E$ then we say that $M$ *satisfies* $E$ and we write

$$M \models E \tag{3}$$

A set of equations $E$ is *consistent* if there exists a well defined feature structure that satisfies $E$.

The notation of the grammar formalism is borrowed from Lexical Functional Grammar [KB82].

**Definition 5** *A* SIMPLE UNIFICATION GRAMMAR $G$ *over a set of attribute symbols $A$ and value symbols $V$ is a 5-tuple $\langle N, T, P, L, S \rangle$ where*

$N$ *is a finite set of symbols, called nonterminals,*

$T$ *is a finite set of symbols, called terminals,*

$P$ is a finite set of production rules

$$A_0 \rightarrow \begin{array}{ccc} A_1 & ... & A_n \\ E_1 & & E_n \end{array} \quad (4)$$

where $n \geq 1$, $A_0, ..., A_n \in N$, and for all $i$, $1 \leq i \leq n$, $E_i$ is a finite set with equations on the forms

$$\uparrow\!\downarrow \psi \doteq \uparrow\!\downarrow \psi' \quad (5)$$
$$\uparrow\!\downarrow \psi'' \doteq v \quad (6)$$

where $\psi, \psi' \in A^*$, $\psi'' \in A^+$ and $v \in V$.[2]

$L$ is a finite set of lexicon rules

$$A \rightarrow \begin{array}{c} t \\ E \end{array} \quad (7)$$

where $A \in N$, $t \in (T \cup \{\varepsilon\})$, and $E$ is a finite set of equations on the form

$$\uparrow\!\downarrow \psi'' \doteq v \quad (8)$$

where $\psi'' \in A^+$ and $v \in V$.

$S$ is a symbol in $N$, called start symbol.

As an example (9) is a production rule.

$$A \rightarrow \begin{array}{ccc} B & C & C \\ \uparrow\doteq\downarrow & \uparrow\doteq\downarrow a_1 & \downarrow\doteq\downarrow a_3\, a_4 \\ \uparrow c \doteq v_1 & \uparrow a_2\, a_3 \doteq\downarrow a_1 & \uparrow a_3 \doteq v_2 \end{array} \quad (9)$$

**Definition 6** A CONSTITUENT STRUCTURE (c-structure) based on a simple unification grammar $G = \langle N, T, P, L, S \rangle$ is a triple $\langle D, C_\mathcal{U}, E_\mathcal{U} \rangle$ where

$D$ is a finite tree domain,

$C_\mathcal{U} : D \rightarrow (N \cup T \cup \{\varepsilon\})$ is a function,

$E_\mathcal{U} : (D - \{\varepsilon\}) \rightarrow \Gamma$ is a function where $\Gamma$ is the set of all equation sets in $P$ and $L$,

such that $C_\mathcal{U}(x) \in (T \cup \{\varepsilon\})$ for all $x \in term(D)$, $C_\mathcal{U}(\varepsilon) = S$, and for all $x \in (D - term(D))$, if $d(x) = n$ then

$$C_\mathcal{U}(x) \rightarrow \begin{array}{ccc} C_\mathcal{U}(x1) & ... & C_\mathcal{U}(xn) \\ E_\mathcal{U}(x1) & & E_\mathcal{U}(xn) \end{array} \quad (10)$$

is a production or lexicon rule in $G$.

The TERMINAL STRING of a constituent structure is the string $C_\mathcal{U}(x_1)...C_\mathcal{U}(x_n)$ where $\{x_1, ..., x_n\} = term(D)$ and $x_i \prec x_{i+1}$ for all $i$, $1 \leq i < n$.

To get equations that can be satisfied by a feature structure we must instantiate the up and down arrows in the equations from the rule set. We substitute them with nodes from the c-structure such that the nodes become the domain of the name mapping. For this purpose we define the '-function such that $E'_\mathcal{U}(xi) = E_\mathcal{U}(xi)[x/\uparrow, xi/\downarrow]$. We see that the value of the function $E'_\mathcal{U}$ is a set of equations that feature structures may satisfy.

---
[2] $\uparrow\!\downarrow$ denotes here a $\uparrow$ or a $\downarrow$

**Definition 7** *The c-structure $\langle D, K, E \rangle$ GENERATES the feature structure $M$ if and only if*

$$M \models \bigcup_{x \in D} E'_{\mathcal{U}}(x) \tag{11}$$

A c-structure may generate different feature structures. The tree domain will form a name set for feature structures that this union generates. A string is grammatical if this union is consistent.

**Definition 8** *A string $w$ is GRAMMATICAL with respect to a simple unification grammar $G$ if and only if there exists a c-structure based on $G$ with $w$ as the terminal string and the c-structure generates a well defined feature structure. The language generated by $G$, $L(G)$ is the set of all grammatical strings with respect to $G$.*

## 4 From Indexed Grammars to Unification Grammars

We are here going to define a simple unification grammar that is equivalent to a given indexed grammar. The main idea is that we use feature structures to represent the index string more or less like a (nested) stack. The use of feature structures to represent stacks for indexed grammars is also used by Gazdar and Mellish [GM89] although they do not go into much details. Here we define a function that transforms any indexed grammar on reduced form with a marked index-end to a simple unification grammar, such that the new grammar generates the same language.

**Definition 9** *Let $G_\$ = \langle N, T, I, P, S \rangle$ be an indexed grammar on reduced form with a marked index-end. We then define the simple unification grammar $\mathcal{U}(G_\$)$ as $\langle N, T, P', L', S \rangle$ where $P'$ and $L'$ are the least sets where*

a) *For each rule on the form $A \to Bf$ in $P$, $P'$ has a production rule on the form*

$$\begin{array}{rl} A \to & B \\ & \downarrow next \doteq \uparrow \\ & \downarrow idx \doteq f \end{array} \tag{12}$$

b) *For each rule on the form $Af \to B$ in $P$, $P'$ has a production rule on the form*

$$\begin{array}{rl} A \to & B \\ & \uparrow next \doteq \downarrow \\ & \uparrow idx \doteq f \end{array} \tag{13}$$

c) *For each rule on the form $A \to BC$ in $P$, $P'$ has a production rule on the form*

$$\begin{array}{rl} A \to & B \quad C \\ & \uparrow \doteq \downarrow \quad \uparrow \doteq \downarrow \end{array} \tag{14}$$

d) For each rule on the form $A \to a$ in $P$, $L'$ has a lexicon rule on the form

$$A \to a \atop \emptyset \qquad (15)$$

If $p$ is a production rule in $G_\$$ then $\mathcal{U}(p)$ is the production or lexicon rule in $\mathcal{U}(G_\$)$ defined by a), b) c) or d).

Notice that there is a one-to-one relation between the production rules in $G_\$$, and production/lexicon-rules in $\mathcal{U}(G_\$)$. We will later define a class of unification grammars which can be defined by production and lexicon rules on the forms used here. But first we will show that $G_\$$ and $\mathcal{U}(G_\$)$ are equivalent.

**Lemma 2** *For every indexed grammar $G_\$$ on reduced form with a marked index end, $L(G_\$) = L(\mathcal{U}(G_\$))$.*

**Proof:** We have to show that for any string $w$, $w \in L(G_\$)$ if and only if $w \in L(\mathcal{U}(G_\$))$.

($\Longrightarrow$) For every $w \in L(G_\$)$ there exists a derivation tree $\langle D, C_\mathcal{I} \rangle$ for $w$ based on $G_\$$. We have to show that based on $\mathcal{U}(G_\$)$ there exist c-structure with $w$ as the terminal string which generates a well defined feature structure. We define the c-structure $\langle D, C_\mathcal{U}, E_\mathcal{U} \rangle$ on the same tree domain $D$.

For every nonterminal node $x$ in $D$ we have a unique production rule $license(x)$ in the indexed grammar, and for each production rule in the indexed grammar we have a unique corresponding production or lexicon rule $\mathcal{U}(license(x))$ in $\mathcal{U}(G_\$)$ according to Definition 9. If

$$\mathcal{U}(license(x)) = A_0 \to A_1 \ldots A_n \atop E_1 \quad\; E_n \qquad (16)$$

then let $C_\mathcal{U}(xi) = A_i$ and $E_\mathcal{U}(xi) = E_i$ for all $1 \leq i \leq n$, and let $C_\mathcal{U}(x) = A_0$. Then we have a valid c-structure and since $C_\mathcal{U}(x) = C_\mathcal{I}^{sym}(x)$ for all $x \in D$, it also has $w$ as terminal string. Now we only have to show that all the equations in the c-structure are satisfied by a well defined feature structure.

For any finite string $\gamma$ over an alphabet $I$ we may define a feature structure where the node set is the union of all suffixes of $\gamma$ and all symbols occurring in $\gamma$. Here we make a distinction between the singleton string of a symbol, and the symbol itself, such that they are regarded as two distinct nodes. For all non-empty string nodes, let the *idx* attribute point to the first symbol of the string and let the *next* attribute point to the rest of the string when we remove the first symbol, ie. $\delta(f\gamma', idx) = f$ and $\delta(f\gamma', next) = \gamma'$ for every non-empty suffix $f\gamma'$ of $\gamma$ where $f \in I$. Let also the atomic value of each symbol-node be the symbol itself, ie. $\alpha(f) = f$. Else, let no more attributes or atomic values be defined, and in particular let $\delta(\varepsilon, next)$, $\delta(\varepsilon, idx)$ and $\alpha(\varepsilon)$ be undefined. We extend the definition directly to any finite set of strings over an alphabet. With any name-mapping to the string nodes defined from this finite set, this is a well defined feature structure since each nonempty string has a unique first symbol, and a unique suffix with length one less than the string itself.

Let $M$ be the feature structure defined as described on the set of all index strings that occur in the derivation tree $\langle D, C_\mathcal{I} \rangle$, with the mapping of each nonterminal node in the tree domain to the index-string of that node: $m(x) = C_\mathcal{I}^{idx}(x)$. This is a well defined feature structure. We now have to show that all the equations in the c-structure are satisfied by the feature structure $M$. We have three different cases to consider:



Assume for a node $x$ that $C_{\mathcal{I}}(x) = A\gamma$ where $\gamma$ is an index-string and that $license(x) = A \to Bf$. Then $C_{\mathcal{I}}(x1) = Bf\gamma$, $m(x) = \gamma$ and $m(x1) = f\gamma$. From $\mathcal{U}(license(x))$ we have that $E'_{\mathcal{U}}(x1) = \{x1\ next \doteq x,\ x1\ idx \doteq f\}$, which is satisfied by the feature structure $M$ since $\delta(f\gamma, next) = \gamma$, and $\alpha(\delta(f\gamma, idx)) = f$.

Assume for a node $x$ that $C_{\mathcal{I}}(x) = Af\gamma$ where $f\gamma$ is an nonempty index-string and that $license(x) = Af \to B$. Then $C_{\mathcal{I}}(x1) = B\gamma$, $m(x) = f\gamma$ and $m(x1) = \gamma$. From $\mathcal{U}(license(x))$ we have that $E'_{\mathcal{U}}(x1) = \{x\ next \doteq x1,\ x\ idx \doteq f\}$, which is satisfied by the index-string feature structure $M$ since $\delta(f\gamma, next) = \gamma$, and $\alpha(\delta(f\gamma, idx)) = f$.

Assume for a node $x$ that $C_{\mathcal{I}}(x) = A\gamma$ where $\gamma$ is an index-string and that $license(x) = A \to BC$. Then $C_{\mathcal{I}}(x1) = B\gamma$, $C_{\mathcal{I}}(x2) = C\gamma$ and $m(x) = m(x1) = m(x2) = \gamma$. From $\mathcal{U}(license(x))$ we have that $E'_{\mathcal{U}}(x1) = \{x \doteq x1\}$ and $E'_{\mathcal{U}}(x2) = \{x \doteq x2\}$, which is satisfied by the index-string feature structure $M$.

We do not have to consider the nodes which license production rules with terminal symbols since all the terminal nodes have empty equation sets. Then all the equations in the c-structure are satisfied by the feature structure $M$ and then $w \in L(\mathcal{U}(G_\$))$.

($\Longleftarrow$) We will here use the function $idx\text{-}lst : Q \to V^*$ defined on any well defined acyclic feature structure as follows: $idx\text{-}lst(q) = \alpha(q)$ if $\alpha(q)$ is defined. If $\delta(q, idx)$ and $\delta(q, next)$ are both defined then $idx\text{-}lst(q)$ is the concatenation of $idx\text{-}lst(\delta(q, idx))$ followed by $idx\text{-}lst(\delta(q, next))$. Else $idx\text{-}lst(q) = \varepsilon$. We restrict our attention to its prefix with \$ as last symbol: Let $idx\text{-}lst_\$ : Q \to V^*$ be the function such that: $idx\text{-}lst_\$(q)$ is the smallest prefix of $idx\text{-}lst(q)$ with \$ as the last symbol. If $idx\text{-}lst(q)$ does not contain any \$ then $idx\text{-}lst_\$(q) = \varepsilon$.

For every $w \in L(\mathcal{U}(G_\$))$ there exists a c-structure $\langle D, C_{\mathcal{U}}, E_{\mathcal{U}}\rangle$ for $w$ based on $\mathcal{U}(G_\$)$ which generates a well defined feature structure. We define the derivation tree $\langle D, C_{\mathcal{I}}\rangle$ for $w$ based on $G_\$$ on the same tree domain $D$. Let $C_{\mathcal{I}}^{sym}(x) = C_{\mathcal{U}}(x)$ for all nodes in $D$ and $C_{\mathcal{I}}^{idx}(x) = idx\text{-}lst_\$(m(x))$ for all nonterminal nodes in $D$ except for the root $\varepsilon$ for which we define $C_{\mathcal{I}}^{idx}(\varepsilon)$ to be the empty string. This derivation tree has $w$ as terminal string, and we just have to show that this is a valid derivation tree according to Definition 2.

Since $G_\$$ has a marked index-end, the only production rule where the start symbol occurs is $S \to A\$$, for an $A \in N$. This gives the following corresponding production rule in $\mathcal{U}(G_\$)$:

$$S \to \begin{array}{c} A \\ \downarrow next \doteq \uparrow \\ \downarrow idx \doteq \$ \end{array} \qquad (17)$$

which is the only production rule in $\mathcal{U}(G_\$)$ where the start symbol occurs. Then $C_{\mathcal{I}}(\varepsilon) = S$ which is the start symbol of $G_\$$. Here we also have that $idx\text{-}lst_\$(m(1)) = \$$ and $C_{\mathcal{U}}(1) = A$ so that $C_{\mathcal{I}}(1) = A\$$ and $S \to A\$$ licenses the root node. For all the other nonterminal nodes in the tree domain we have four cases to consider:

Assume for a nonterminal node $x$ except for the root node that $C_{\mathcal{U}}(x) = A$ and $idx\text{-}lst_\$(m(x)) = \gamma$. Then $C_{\mathcal{I}}(x) = A\gamma$. Assume also that there exists a production rule in $\mathcal{U}(G_\$)$ from Definition 9 a), such that $C_{\mathcal{U}}(x1) = B$, $E'_{\mathcal{U}}(x1) = \{x1\ next \doteq x, x1\ idx \doteq f\}$ and $x1$ has no sister nodes. Since \$ only occurs in the one production rule with the start symbol, $f \neq \$$. Then $idx\text{-}lst_\$(m(x1)) = f\gamma$ and $C_{\mathcal{I}}(x1) = Bf\gamma$. From the reverse of Definition 9 a), there exists a production rule $A \to Bf$ in $G_\$$, which licenses $x$.

Assume for a nonterminal node $x$ except for the root node that $C_{\mathcal{U}}(x) = A$ and $idx\text{-}lst_\$(m(x)) = f\gamma$. Then $C_{\mathcal{I}}(x) = Af\gamma$. Assume also that there exists a production rule in $\mathcal{U}(G_\$)$ from Definition 9 b), such that $C_{\mathcal{U}}(x1) = B$, $E'_{\mathcal{U}}(x1) = \{x\ next \doteq x1,\ x\ idx \doteq f\}$ and $x1$ has no sister nodes. Since \$ only occur in the

one production rule with the start symbol, $f \neq \$$. Then $idx\text{-}lst_\$(m(x1)) = \gamma$ and $C_\mathcal{I}(x1) = B\gamma$. By the reverse of Definition 9 b), there exist a production rule $Af \to B$ in $G_\$$, which licenses $x$.

Assume for a nonterminal node $x$ except for the root node that $C_\mathcal{U}(x) = A$ and $idx\text{-}lst_\$(m(x)) = \gamma$. Then $C_\mathcal{I}(x) = A\gamma$. Assume also that there exist a production rule in $\mathcal{U}(G_\$)$ from Definition 9 c), such that $d(x) = 2$, $C_\mathcal{U}(x1) = B$, $C_\mathcal{U}(x2) = C$, $E'_\mathcal{U}(x1) = \{x \doteq x1\}$ and $E'_\mathcal{U}(x2) = \{x \doteq x2\}$. Then $idx\text{-}lst_\$(m(x1)) = idx\text{-}lst_\$(m(x2)) = \gamma$, $C_\mathcal{I}(x1) = B\gamma$ and $C_\mathcal{I}(x2) = C\gamma$ By the reverse of Definition 9 c), there exist a production rule $A \to BC$ in $G_\$$, which licenses $x$.

Assume for a nonterminal node $x$ except for the root node that $C_\mathcal{U}(x) = A$ and $idx\text{-}lst_\$(m(x)) = \gamma$. Then $C_\mathcal{I}(x) = A\gamma$. Assume also that there exists a lexicon rule in $\mathcal{U}(G_\$)$ from Definition 9 d), such that $d(x) = 1$, $C_\mathcal{U}(x1) = t$ and $E'_\mathcal{U}(x1) = \emptyset$. Then $C_\mathcal{I}(x1) = t$. By the reverse of Definition 9 d), there exist a production rule $A \to t$ in $G_\$$ which licenses $x$.

We then have a valid derivation tree with the same terminal string as the c-structure and then $w \in L(G_\$)$. □

**Example 2** Let $G = \langle N, T, I, P, S \rangle$ be an indexed grammar where $T = \{d\}$ is the set of terminal symbols, $N = \{S, A, B, C, C', D\}$ is the set of nonterminal symbols, $I = \{\$, f, g\}$ is the set of indices and $P$ is the least set containing the following production rules:

$$
\begin{array}{lll}
S \to A\$ & B \to CC & \\
A \to Bf & Cg \to C' & C' \to CC \\
B \to Bg & Cf \to D & D \to d
\end{array}
$$

This grammar is on reduced form with a marked index-end. The simple unification grammar $\mathcal{U}(G)$ as given in Definition 9 is then the 5-tuple $\langle N, T, P', L', S \rangle$ where $P'$ is the least set containing the following production rules:

$$
\begin{array}{lll}
S \to \begin{array}{c} A \\ \downarrow next \doteq \uparrow \\ \downarrow idx \doteq \$ \end{array} & B \to \begin{array}{cc} C & C \\ \uparrow \doteq \downarrow & \uparrow \doteq \downarrow \end{array} & \\
\\
A \to \begin{array}{c} B \\ \downarrow next \doteq \uparrow \\ \downarrow idx \doteq f \end{array} & C \to \begin{array}{c} C' \\ \uparrow next \doteq \downarrow \\ \uparrow idx \doteq g \end{array} & C' \to \begin{array}{cc} C & C \\ \uparrow \doteq \downarrow & \uparrow \doteq \downarrow \end{array} \\
\\
B \to \begin{array}{c} B \\ \downarrow next \doteq \uparrow \\ \downarrow idx \doteq g \end{array} & C \to \begin{array}{c} D \\ \uparrow next \doteq \downarrow \\ \uparrow idx \doteq f \end{array} &
\end{array}
$$

and $L'$ contains one single lexicon rule:

$$
\begin{array}{c}
D \to d \\
\emptyset
\end{array}
$$

Figure 2 shows the derivation tree for the string *"dddd"* based on the indexed grammar $G$ together with the c-structure and the feature structure for the same string string based on the simple unification grammar $\mathcal{U}(G)$. This shows that the string *"dddd"* is both in $L(G)$ and in $L(\mathcal{U}(G))$. The language generated by $G$ and $\mathcal{U}(G)$ is $\{d^{2^n} \mid n \geq 1\}$.

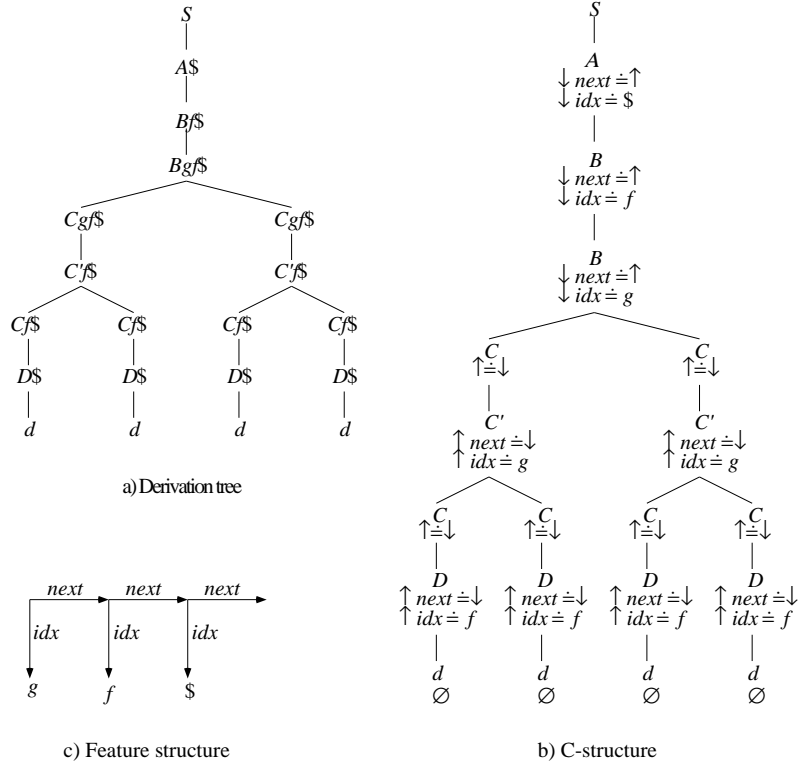

Figure 2: *Derivation tree (a) for the string "dddd" based on the grammar G in Example 2, together with the c-structure (b) and feature structure (c) for the same string based on the grammar $\mathcal{U}(G)$.*

## 5   A Unification Grammar Formalism for Indexed Languages

We are here going to define a version of the simple unification grammar that describes the class of indexed languages. Just to be precise, a *class of languages*, $\mathcal{C}_\Gamma$ over a countable set $\Gamma$ of symbols is a set of languages, such that each language $L \in \mathcal{C}_\Gamma$ is a subset of $\Sigma^*$ where $\Sigma$ is a finite subset of $\Gamma$. The class $\mathcal{C}_\Gamma(GF)$ of languages that a grammar formalism $GF$ describes is the set of all languages $L'$ over $\Gamma$ such that there exists a grammar $G$ in $GF$ where $L(G) = L'$. The class of indexed languages is then the set of languages such that there for each language exist a indexed grammar that generates the language. We assume that $\Gamma$ is the set of all terminal symbols that we use and drop $\Gamma$ as subscript.

**Definition 10** *A* UNIFICATION GRAMMAR FOR INDEXED LANGUAGES, $\mathcal{UGI}$ *is a simple unification grammar where*

  a) *each equation set in the production rules is on one of the three forms*

   - $E = \{\uparrow \doteq \downarrow\}$,
   - $E = \{\downarrow\ next \doteq \uparrow, \downarrow\ idx \doteq f\}$,
   - $E = \{\uparrow\ next \doteq \downarrow, \uparrow\ idx \doteq f\}$

   *where $f$ is any value symbol, and next and idx are the same two attribute symbols for all equations in all production rules in $\mathcal{UGI}$,*

*b) each lexicon rule has en empty equation set.*

**Lemma 3** *The class of languages $\mathcal{C}(\mathcal{UGI})$ contains the class of indexed languages.*

**Proof:** Aho [Aho68] showed that for every indexed language there exists an indexed grammar on reduced form which generates the language. From Lemma 1 and its proof we have that for every indexed grammar $G$ on reduced form there exists an indexed grammar on reduced form with a marked index-end $G_\$$, such that $L(G) = L(G_\$)$. The simple unification grammar $\mathcal{U}(G_\$)$ defined from the indexed grammar on reduced form with a marked index-end in Definition 9 is an $\mathcal{UGI}$ grammar. From Lemma 2 we have that $L(G_\$) = L(\mathcal{U}(G_\$))$. Then every indexed language can be generated by an $\mathcal{UGI}$ grammar. □

We shall now show that every $\mathcal{UGI}$ grammar generates an indexed language, but to do this we need some technical results. First it is easy to see that every $\mathcal{UGI}$ grammar can be formulated with rules only on the forms used in Definition 9 *a)-d)*. We define the reduced form for this.

**Definition 11** *A $\mathcal{UGI}$ grammar is on* REDUCED FORM *if and only if every production rule is on one of the three following forms:*

$$
\begin{array}{cccccccc}
A & \rightarrow & B & & A & \rightarrow & B & & A & \rightarrow & B & C \\
& & \downarrow next \doteq \uparrow & & & & \uparrow next \doteq \downarrow & & & & \uparrow \doteq \downarrow & \uparrow \doteq \downarrow \\
& & \downarrow idx \doteq f & & & & \uparrow idx \doteq f
\end{array} \quad (18)
$$

**Lemma 4** *For every $\mathcal{UGI}$ grammar there is an equivalent grammar on reduced form.*

**Proof:** Using the techniques from the standard proof for normal form for context-free grammars, it is straight forward to replace each production rule in the original grammar not on reduced form with a set of new lexicon rules and production rules on reduced form. This can be done such that one instance of an original rule corresponds to the net effect of combining one ore more of the new rules. This is possible since we allow the empty string in lexicon rules. □

To make this formalism more directly comparable to indexed grammars with a marked index-end we use what we will call a *sink-mapped root*:

**Definition 12** *A $\mathcal{UGI}$ grammar $\langle N, T, P, L, S \rangle$ has a* SINK-MAPPED ROOT *if and only if it has one and only one production rule where the start symbol occurs and this rule is on the form*

$$
\begin{array}{ccc}
S & \rightarrow & A \\
& & \downarrow next \doteq \uparrow \\
& & \downarrow idx \doteq \$
\end{array} \quad (19)
$$

*where $A \in N$ and the value symbol $\$$ does not occur in any other production rule.*

The value symbol $\$$ will form some kind of a blockade in the feature structure since it does not occur in any other production rule, hence no other node in the c-structure will be mapped to the same node in the feature structure as the root of the c-structure.

What we are doing here is to put a mark at the bottom of the stack of indices, in the way the nested stack is represented as a feature structure. We also want to map the root of the c-structure to the "sink" of the feature structure when we follow the *next* attribute.

**Lemma 5** *For every $\mathcal{UGI}$ grammar $G$ there exists a $\mathcal{UGI}$ grammar with a sink-mapped root $G'$ such that $L(G) = L(G')$.*



**Proof:** First we show how we from any $\mathcal{UGI}$ grammar $G$ may define a $\mathcal{UGI}$ grammar with a sink-mapped root $G'$. After this we show that for any string $w$, $w \in L(G)$ if and only if $w \in L(G')$.

Let any $\mathcal{UGI}$ grammar $G = \langle N, T, P, L, S \rangle$ be given, and assume that $S_0$, $S'$ and $S_\varepsilon$ are neither terminal nor nonterminal symbols in $G$, and that \$ is a value symbol not used in $G$. The grammar $G'$ is defined by adding the following production and lexicon rules to the rules we have in $G$:

*i*) Let the following be two production rules:

$$S_0 \rightarrow \begin{array}{c} S' \\ \downarrow next \doteq \uparrow \\ \downarrow idx \doteq \$ \end{array} \tag{20}$$

$$S' \rightarrow \begin{array}{cc} S & S_\varepsilon \\ \uparrow \doteq \downarrow & \uparrow \doteq \downarrow \end{array} \tag{21}$$

*ii*) For each $f \in V$ used in any production rule in $G$, let the following be a production rule:

$$S' \rightarrow \begin{array}{c} S' \\ \downarrow next \doteq \uparrow \\ \downarrow idx \doteq f \end{array} \tag{22}$$

*iii*) Let the following be a lexicon rule:

$$S_\varepsilon \rightarrow \begin{array}{c} \varepsilon \\ \emptyset \end{array} \tag{23}$$

Complete $G'$ by adding $S_0$, $S'$ and $S_\varepsilon$ to the nonterminal symbols, and let $S_0$ be the start symbol of $G'$. We see that $G'$ is a $\mathcal{UGI}$ grammar with a SINK-MAPPED ROOT. Notice also that if $G$ is on reduced form so is the new grammar.[3]

Now we have to show that for any string $w$, $w \in L(G)$ if and only if $w \in L(G')$.

($\Longrightarrow$) We show this direction in two steps: First we define something that we call a *canonical* feature structure for c-structures based on $\mathcal{UGI}$ grammars. This is done such that if the c-structure generates a well defined feature structure at all, then it is also generating the canonical feature structure. After this definition we show how we from a c-structure based on $G$, together with its canonical feature structure can construct a c-structure together with a feature structure based on the grammar $G'$. This is done such that the two c-structures have the same terminal string and if the terminal string is in $L(G)$ so is it in $L(G')$ also.

Let $\langle D, K, E \rangle$ be any c-structure based on a $\mathcal{UGI}$ grammar $G$ such that it generates a feature structure. The *canonical feature structure* $\langle Q, \delta, \alpha, m \rangle$ for the c-structure is defined as follows: Let first $Q_+$ be the set of all sequences of nodes from the c-structure with at most $2n + 1$ nodes in each sequence, where $n$ is the height of the c-structure. Then let the name mapping function $m$ be defined on $Q_+$ by top-down induction on the nodes in the c-structure: First let the mapping of the root node, $m(\varepsilon)$ be the sequence of $n + 1$ $\varepsilon$'s, $<\varepsilon, \varepsilon, ..., \varepsilon>$, where again $n$ is

---
[3] The use of $S_\varepsilon$ in rule (21) together with rule (23) where it will label the mother of a node with the empty string is only done because we want to stay in the domain of grammars on reduced form when $G$ is on reduced form. This definition could be simplified if we did not want this.



the height of the c-structure. Now assume that $m(x)$ is defined for a node $x$ in the c-structure. Then for each daughter $xi$ of $x$, let

$$\begin{array}{lll} m(xi) = m(x) & \text{if} & \uparrow \doteq \downarrow \in E'(xi) \\ m(xi) = pop(m(x)) & \text{if} & \uparrow next \doteq \downarrow \in E'(xi) \\ m(xi) = add(xi, m(x)) & \text{if} & \downarrow next \doteq \uparrow \in E'(xi) \end{array} \quad (24)$$

where $pop$ of any nonempty sequence is the sequence we get by removing the first element, $pop(<x_1, x_2, ..., x_k>) = <x_2, ..., x_k>$, and $add$ of a single element and a sequence is the sequence we get by adding the single element to the beginning of the sequence, $add(x, <x_1, ..., x_k>) = <x, x_1, ..., x_k>$. Since the root node is mapped to the sequence of $n+1$ $\varepsilon$'s, $pop$ and $add$ may not go out of their domain and therefore is $m$ well defined.

Extend now the set $Q_+$ such that all the value symbol used in the c-structure also are elements in $Q_+$. Then let the partial function $\delta_+ : Q_+ \times \{next, idx\} \to Q_+$ be defined such that $\delta_+(q, next) = pop(q)$ for all nonempty sequences $q \in Q_+$, and let $\delta_+(q, next)$ be undefined when $q$ is the empty sequence. Moreover let $\delta_+(q, idx) = f$ for the value symbol $f$ if and only if there exists a node $x$ in the c-structure such that either $\downarrow idx \doteq f \in E(x)$, or $\uparrow idx \doteq f \in E(xi)$ for a daughter $xi$ of $x$. This is the only place where inconsistency may occur and we will later see that it will not occur if the c-structure generates any feature structure at all. We extend the definition of the $\delta_+$ to pairs of nodes and strings of the attribute symbols as described in the definition of feature structures in the beginning of section 3.

Now, let us shrink the definitions of $Q_+$ and $\delta_+$ such that we get a well defined feature structure. First let $Q \subseteq Q_+$ be the set of all nodes that is reachable from a named node, formally $Q = \{q \mid \exists x \in D, \psi \in \{next, idx\}^* : \delta_+(m(x), \psi) = q\}$. Then, we restrict $\delta$ to the new domain: $\delta = \delta_+ \cap (Q \times \{next, idx\} \times Q)$. Finally, let $\alpha(f) = f$ for all value symbol used in the c-structure. We now have a feature structure and it is describable and acyclic directly from the definition of $Q$ and $\delta$. It is also atomic since $\delta$ is not defined on any feature symbol node, and $\alpha$ is only defined on feature symbol nodes. Moreover, it satisfies all the equations from the c-structure after we have instantiated the up and down arrows. We will now show that if the c-structure generates any well defined feature structure so will it generate the well defined canonical one also.

Let $M' = \langle Q', \delta', \alpha', m' \rangle$ be any well defined feature structure which the c-structure generates, and assume that we have the canonical feature structure as described. From the fact that the c-structure generates a feature structure, and from the definition of the canonical feature structure we have that if $m(x) = m(y)$ for any two nodes $x$ and $y$ in the c-structure then $m'(x) = m'(y)$. Now we may define a function $h : Q \to Q'$ from the nodes in the canonical feature structure to the nodes in $M'$, such that $m'(x) = h(m(x))$ for all nodes $x$ in the c-structure. Assume then that we don't have a well defined canonical feature structure because of inconsistency in it definition. This means that there exist two instantiated equations, $x \ idx \doteq f$ and $y \ idx \doteq f'$ from the c-structure where $m(x) = m(y)$ but $f \neq f'$. However, then $m'(x) = m'(y)$, and inconsistency must also occur with respect to $M'$ and the c-structure can not generate any well defined feature structure. Then the canonical feature structure must be consistent defined, and since it is also describable, acyclic and atomic it is well defined. Since it also satisfies all the equations in the c-structure it is generated by the c-structure.

Now we have a well defined canonical feature structure for each c-structure based on any $\mathcal{UGI}$ grammar if the c-structure generates a feature structure. Notice that $\delta(<\varepsilon>, idx)$ is not defined for the canonical feature structure. This due to the mapping of the root in the c-structure to the sequence of $n+1$ $\varepsilon$'s, where $n$ is the height of the c-structure. With this height it is only possible to pop of $n-1$ $\varepsilon$'s



according to definition of the name mapping (24), and since $\delta(q, idx)$ is only defined for $q$ if there exist a node $x$ such that $m(x) = q$, $\delta(<\varepsilon>, idx)$ can not be defined.

Assume now that $w \in L(G)$ for a grammar $G$. Then we have a c-structure for $w$ based on $G$ which generates a well defined feature structure. Then it is also generating a canonical feature structure $M = \langle Q, \delta, \alpha, m \rangle$ as described above. For this feature structure we extend the definition of $\delta$ and $\alpha$ as follows: First let $\delta(<\varepsilon>, idx) = \$$ and let $\alpha(\$) = \$$. For all sequences $q$ of $\varepsilon$'s such that $\delta(q, idx)$ is not defined, let $\delta(q, idx) = f$ for any value symbol $f$ which occurs in the c-structure. When we construct the new c-structure based on $G'$ the old nodes keep their mapping values.

We construct a new c-structure for $w$ based on $G'$ by the following steps: First add a new node on the top of the c-structure by applying the production rule (21). This give us also a new sister node for the old root node. Map the two new nodes to the same node in the extended canonical feature structure as the old root node. This secures that the equations in the production rule (21) is satisfied by the extended feature structure. The new sister node labeled with $S_\varepsilon$ may only be a mother of a terminal node labeled with the empty string such that the terminal string is still $w$. Now add $n$ nodes above the present root node by applying the generic production rule (22) $n-1$ times and production rule (20) on the topmost node. This top node will be the root node in the new c-structure and it is now labeled with the start symbol in $G'$. When applying the generic production rule (22), let $f = \alpha(\delta(m(x1), idx))$ for each new node $x$ where it is applied. The new nodes are each mapped to the sequence of $k$ $\varepsilon$'s, where $k$ is the node's distance from the new root node. In this way the new root node is mapped to the empty sequence, the daughter of the root node is mapped to $<\varepsilon>$, and so on. Since $\delta(<\varepsilon>, idx) = \$$ the equations in production rule (20) is satisfied by the feature structure. Moreover since $f = \alpha(\delta(m(x1), idx))$ for each node $x$ where the production rule (22) is applied and $\delta(q, next) = pop(q)$, all the equations is satisfied by the feature structure. We then have a c-structure based on $G'$ with $w$ as terminal string, and this c-structure generates a well defined feature structure. Then $w \in L(G')$.

($\Longleftarrow$) Assume that $w \in L(G')$ for a grammar $G$. Then there is a c-structure with category $S_0$ in the root, and a sequence of derivations down to a node with category $S$, where each intermediate node has category $S'$. This has been constructed by first using production rule (20) and then a sequence of zero or more applications of production rule (22) before production rule (21) gives the node with category $S$. Every node above the first node with category $S$ has only one child, except the first which has an additional daughter, labeled with $S_\varepsilon$. This daughter is the mother of a single terminal node labeled with the empty string. Then we can remove all nodes above the node labeled $S$ and still have the same terminal string $w$ in the c-structure. The new c-structure will have a root-node with category $S$, and only production rules from the grammar $G$ are used. Since the original c-structure generates a feature structure, so does the new one. Then $w \in L(G)$. □

Now we have the necessary technical results to show that every language in $\mathcal{C}(\mathcal{UGI})$ is an indexed language. We do this in two steps.

**Lemma 6** *For any $\mathcal{UGI}$ grammar $G$ on reduced form with a sink-mapped root, there exists an indexed grammar $G_\mathcal{I}$ such that $\mathcal{U}(G_\mathcal{I}) = G$.*

**Proof:** Assume that $G = \langle N, T, P, L, S \rangle$ is a $\mathcal{UGI}$ grammar on reduced form with a sink-mapped root. Then let $G_\mathcal{I} = \langle N, T, I', P', S \rangle$ be an indexed grammar where $I'$ is all the value symbols occurring in $G$, and $P'$ is constructed from $P$ and $L$ by reversing Definition 9 a)-d). This can bee done since $G$ is on reduced form and there exist a one to one relation between the production rules in the indexed grammar and the production and lexicon rules in the unification grammar defined there. Since $G$ has a sink-mapped root the start symbol will occur in one and only one production rule together with a unique value symbol. Then $G_\mathcal{I}$ has a marked index-end and $\mathcal{U}(G_\mathcal{I}) = G$. □

**Lemma 7** *Every language in $\mathcal{C}(\mathcal{UGI})$ is an indexed language.*

**Proof:** From Lemma 4 and Lemma 5 we have for any language in $\mathcal{C}(\mathcal{UGI})$ that there exist a $\mathcal{UGI}$ grammar $G$ on reduced form with a sink-mapped root that generates the language. From Lemma 6 we have an indexed grammar $G_\mathcal{I}$ such that $\mathcal{U}(G_\mathcal{I}) = G$. By Lemma 2 we have that $L(G_\mathcal{I}) = L(G)$. Then we have an indexed grammar for all languages in $\mathcal{C}(\mathcal{UGI})$. □

From Lemma 3 and Lemma 7 we then have the following result:

**Theorem 1** : *The class $\mathcal{C}(\mathcal{UGI})$ is the class of indexed languages.*

# Acknowledgments

I would like to thank Tore Langholm for his advice during the work that this paper is based on, and for extended comments on earlier versions of this paper.

# References

[Aho68] Alfred V. Aho. Indexed grammars —an extension of context-free grammars. *Journal of the Association of Computing Machinery*, 15(4):647–671, October 1968.

[Col91] Erik A. Colban. *Three Studies in Computational Semantics*. Dr.scient thesis, University of Oslo, 1991.

[Gal86] Jean H. Gallier. *Logic for Computer Science*. Harper & Row, Publishers, New York, 1986.

[Gaz88] Gerald Gazdar. Applicability of indexed grammars to natural languages. In Uwe Reye and Christian Rohrer, editors, *Natural Language Parsing and Linguistic Theories*, pages 69–94. D. Reidel Publishing Company, Dordrecht, Holland, 1988.

[GM89] Gerald Gazdar and Chris Mellish. *Natural Language Processing in LISP*. Addison-Wesley Publishing Company, 1989. Also in Prolog version.

[Hay73] Takeshi Hayashi. On derivation trees of indexed grammars —an extension of the uvwxy-teorem—. *Publications of the Research Institute for Mathematical Sciences, Kyoto University*, 9(1):61–92, 1973.

[HU79] John E. Hopcroft and Jeffrey D. Ullman. *Introduction to Automata Theory, Languages and Computation*. Addison-Wesley, 1979.

[KB82] Ronald M. Kaplan and Joan Bresnan. Lexical functional grammar: A formal system of grammatical representation. In Joan Bresnan, editor, *The Mental Representation of Gramatical Relations*. MIT-Press, 1982.

[Shi85] Stuart M. Shieber. Evidence against the context-freeness of natural language. *Linguistics and Philosophy*, 8:333–343, 1985.